\documentclass{article}

\usepackage{arxiv}

\usepackage[utf8]{inputenc} % allow utf-8 input
\usepackage[T1]{fontenc}    % use 8-bit T1 fonts
\usepackage{hyperref}       % hyperlinks
\usepackage{url}            % simple URL typesetting
\usepackage{booktabs}       % professional-quality tables
\usepackage{amsfonts}       % blackboard math symbols
\usepackage{nicefrac}       % compact symbols for 1/2, etc.
\usepackage{microtype}      % microtypography
\usepackage{lipsum}		% Can be removed after putting your text content
\usepackage{graphicx}
\usepackage{natbib}
\usepackage{doi}
\usepackage{tabularx}
\usepackage{subcaption}

\title{Thermal Analysis of Malignant Brain Tumors by Employing a Morphological Differentiation-Based Method in Conjunction with Artificial Neural Network}

\author{ \href{https://orcid.org/0000-0002-6447-717X}{\includegraphics[scale=0.06]{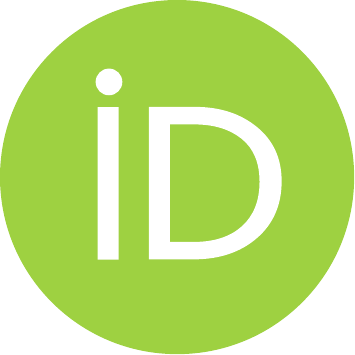}\hspace{1mm}Hamed Hani}\thanks{Spine Research Institute: spine.osu.edu} \\
	Department of Integrated Systems Engineering\\
	The Ohio State University\\
	Columbus, OH 43210 \\
	\texttt{hani.4@osu.edu} \\
	\And
	\href{https://orcid.org/0000-0000-0000-0000}{\includegraphics[scale=0.06]{orcid.pdf}\hspace{1mm}Afsaneh Mojra} \\
	Department of Mechanical Engineering\\
	K.N. Toosi University of Technology\\
	Tehran, Iran \\
	\texttt{mojra@kntu.ac.ir} \\
}

\renewcommand{\shorttitle}{Thermal Analysis of Malignant Brain Tumors}

\hypersetup{
pdftitle={Thermal Analysis of Malignant Brain Tumors by Employing a Morphological Differentiation-Based Method in Conjunction with Artificial Neural Network},
pdfsubject={q-bio.NC, q-bio.QM},
pdfauthor={Hamed Hani, Afsaneh Mojra},
pdfkeywords={Brain tumor, Tumor differentiation, Morphological analysis, Artificial neural network, Finite element method},
}

\begin{document}
\maketitle

\begin{abstract}
	In this study, a morphological differentiation-based method has been introduced which employs temperature distribution on the tissue surface to detect brain tumor’s malignancy. According to the common tumor CT scans, two different scenarios have been implemented to describe irregular shape of the malignant tumor. In the first scenario, tumor has been considered as a polygon base prism and in the second one, it has been considered as a star-shaped base prism. By increasing the number of sides of the polygon or wings of the star, degree of the malignancy has been increased.  Constant heat generation has been considered for the tumor and finite element analysis has been conducted by the ABAQUS software linked with a PYTHON script on both tumor models to study temperature variations on the top tissue surface. This temperature distribution has been characterized by 10 parameters. In each scenario, 98 sets of these parameters has been used as inputs of a radial basis function neural network (RBFNN) and number of sides or wings has been selected to be the output. The RBFNN has been trained to identify malignancy of tumor based on its morphology. According to the RBFNN results, the proposed method has been capable of differentiating between benign and malignant tumors and estimating the degree of malignancy with high accuracy.
\end{abstract}

% keywords can be removed

\section{Introduction}
Brain tumors are categorized into benign and malignant. Despite considerable advances in the diagnosis and the treatment of the brain tumors, the mortality from malignant brain tumors is still high~\cite{Kateb_2009}. Malignant tumors are cancerous and are made up of cells that grow out of control~\cite{JamesW.Baish2000}. It has been found that cells in the peripheral areas of the malignant tumor have a high invasive rhythm to the surrounding tissue~\cite{guarino_2007}. Therefore, border of a malignant tumor has sharp edges that result in a polygonal shape or star-shaped morphology of the tumor~\cite{Golston1992}. Malignant tumors are deeply fixed in the surrounding tissue by the sharp edges. The sharpness increases rapidly since the tumor needs more invasions to the nearby tissue for the growth~\cite{Condeelis_2006}. On the contrary, benign tumors are often smooth and round and easy to remove since they are not attached to the surrounding tissue~\cite{Shah_1995}.\\
Surgery is usually the first step in the treatment of the brain tumors. The goal is to remove as much of the tumor as possible while maintaining neurological function. Surgical treatments for malignant brain tumors are not easy to handle because they do not have clear borders~\cite{Argani_2001}. In order to improve the surgeon’s ability in defining the tumor’s border and avoid injury to vital brain areas in the operating room, image-guided surgery is performed. Intraoperative imaging is a revolutionary tool in the modern neurosurgery~\cite{Proctor2005}. Intraoperative imaging techniques especially intraoperative MRI (iMRI), help neurosurgeons achieve the goal of maximum tumor resection with least morbidity~\cite{Schulder2003}. The main drawbacks of using iMRI is the patient positioning during the surgery for a proper imaging and also limitation of using the surgical instruments because of the presence of a strong magnetic field. \\
In order to avoid limitations and high expenses of intraoperative imaging, many researches have focused on improving the performance of the preoperative tumor imaging techniques. These techniques mainly include magnetic resonance imaging (MRI) and computed tomography scan (CT scan). In the MRI, an injected contrast agent is used which makes the cancerous tumor brighter than the surrounding normal tissue. During the scan, there is a rapid increase in the signal intensity of a malignant tumor immediately 1 to 2 minutes after the injection. The intensity decreases in the following minutes\cite{KOBAYASHI_2005}. For a benign mass, the rise in the intensity is much slower. Inaccuracy in the time and the intensity measurements results in a probability that benign and malignant tumors have overlap in their morphological appearance~\cite{Barentsz_1996}. It was proved that the specificity of MRI to correctly predict a benign tumor is limited. Moreover, Specificity of MRI would be decreased by reducing the tumor size. Therefore, MRI is usually recommended after a malignancy is detected by other methods in order to have more information about the extent of the cancer.\\
CAT or CT scanning is an accurate medical test that combines x-ray with computerized technology to detect malignancy. The main drawback of this method is using high doses of radiation, leading to the possibility of lung cancer or breast cancer as a consequence~\cite{Lee_2004}. X-rays also damage DNA itself~\cite{Spotheim_Maurizot_2011}. CT scanning provides images in shades of grey; occasionally the shades are similar, making it difficult to distinguish between the normal and abnormal tissues. To overcome such deficiency a contrast agent may be injected into the bloodstream. Main problems of the injection include pathological side-effects such as nausea and vomiting, hypotension and extravasation of the contrast which can be severe enough to require skin grafting~\cite{Rull2015}. \\
During recent years, many researches have focused on improving the procedure of defining tumor morphology in the preoperative imaging techniques. \citet{Wu_2012} used the level set method to segment ultrasound breast tumors automatically and used a genetic algorithm to detect indicative features for the support vector machine (SVM) to detect tumor malignancy. The proposed system could discriminate benign from malignant breast tumors with high accuracy and short feature extraction time. \citet{Huang_2013} evaluated the value of using 3D breast MRI morphological features to differentiate malignant and benign breast tumor. Malignancy of a tumor was investigated by using a number of extracted morphological features in the breast MRI. \citet{Jen_2015} introduced a method for abnormality detection in the mammograms based on an abnormality detection classifier (ADC) which extracts a couple of distinctive features including first-order statistical intensities and gradients. In this method, image preprocessing techniques were used to obtain more accurate breast tissue segmentation. \citet{Han_2015} provided an improved segmentation algorithm which is a combination of the fuzzy clustering segmentation and the fuzzy edge enhancement. Results showed that the fuzzy clustering segmentation is highly efficient for complex brain tissues, and the images after the fuzzy clustering segmentation provide a solid foundation for 3D processing and help to acquire better 3D visualization of the brain tumors. \citet{Ramya_2015} developed a robust segmentation  algorithm  in order to diagnose  tumors in the MR images. In this method, a 4th-order partial differential equation was employed to denoise images to improve segmentation accuracy. \citet{Zhang_2015} proposed a wavelet energy-based method to classify the MR images. This approach had a three-stage system which detects characteristics indicative of the abnormal brain tissues. \citet{Shirazi_2016} used a combination of support vector machine (SVM) and mixed gravitational search algorithm (MGSA) to improve the classification accuracy in the mammography images. \citet{Xia_2016} proposed a novel voting ranking random forests (VRRF) method for the image classification and developed a center-proliferation segmentation (CPS) method. This method showed good performance in the image classification with strong robustness.\\
In the present study, a palpation-based method was used for scanning of the brain tissue in order to detect and follow the malignancies. The method avoids the main aforementioned drawbacks of the imaging techniques since it is only based on the tissue palpation. It is called “tactile thermography” and maps the thermal parameters of the tissue mainly the temperature and the heat flux on the tissue surface. \citet{Sadeghi_Goughari_2015} estimated thermal parameters of the brain tumors by introducing the tactile thermography method as a new noninvasive thermal imaging method. In this method, the brain tissue was mechanically and thermally loaded and the temperature and the heat flux variations on the tissue surfaces were recorded. A number of thermal parameters were extracted and optimized by an artificial neural network to verify tumor existence and its depth. The main objective of this study is to evaluate the capability of the proposed method in detecting sharp morphology of the tumor indicative of the tumor malignancy and also to verify its sensitivity to the sharpness increase that can be used in the follow-up procedure for evaluating the tumor growth.\\
To this end, a malignant tumor is simulated in the brain tissue by two scenarios that represent the invading sharp edges of the tumor. Moreover, the tumor is considered as a heat source in the thermal analysis since the cell number and the overall cell metabolism are considerably increased in the tumor relative to the normal tissue. By conducting a thermal analysis, the temperature distribution on the tissue surface is obtained for tumors with varying penetration into the surrounding tissue. By the use of the artificial neural network, a number of variables are extracted from the temperature map and will be used as the malignancy indicative features.

\section{Materials and Methods}
\begin{figure}
	\centering
    \includegraphics[width=0.5\textwidth]{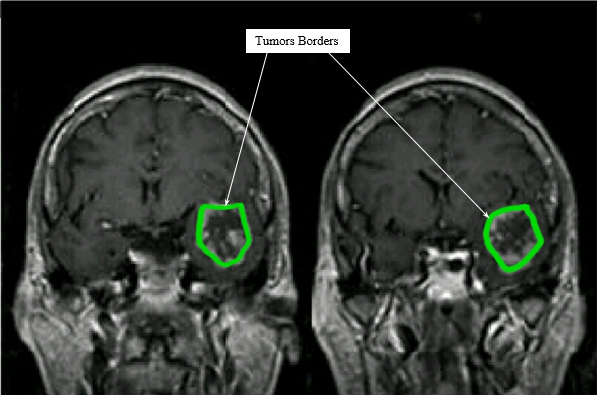}
	\caption{CT scan image of a malignant tumor resembles a polygon presented as scenario 1.}
	\label{fig:ct}
\end{figure}

\begin{figure}
	\centering
    \includegraphics[width=0.5\textwidth]{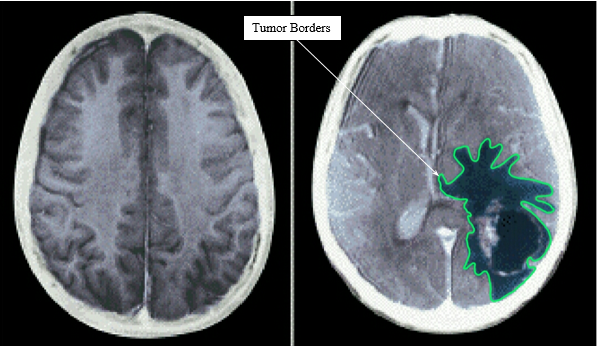}
	\caption{CT scan image of a malignant tumor resembles a star presented as scenario 2.}
	\label{fig:ct2}
\end{figure}

\begin{figure}
	\centering
    \begin{subfigure}{0.5\textwidth}
        \includegraphics[width=\textwidth]{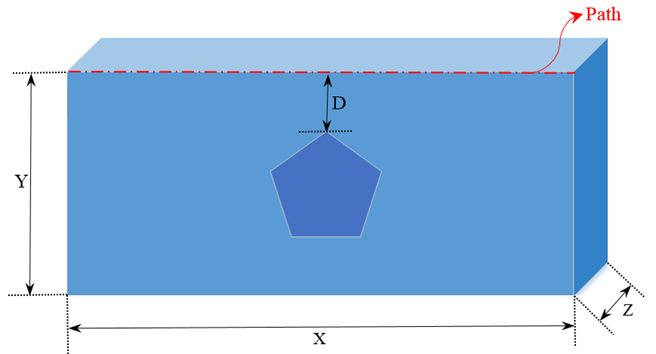}
        \caption{}
        \label{fig:prismmodelfirst}
    \end{subfigure}
    \begin{subfigure}{0.5\textwidth}
        \includegraphics[width=\textwidth]{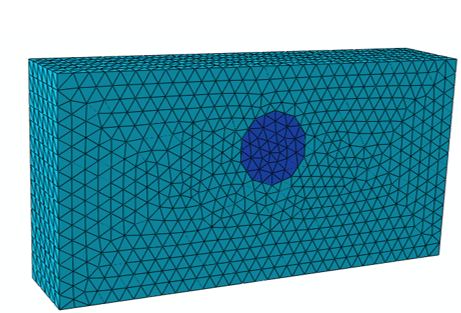}
        \caption{}
        \label{fig:prismmodelsecond}
    \end{subfigure}
	\caption{Mid cross section of the brain tissue model (rectangular cuboid) in the ABAQUS environment including (a) a pentagonal based prismatic tumor; (b) a decagonal based prismatic tumor.}
	\label{fig:prismmodel}
\end{figure}

\begin{figure}
	\centering
    \begin{subfigure}{0.5\textwidth}
        \includegraphics[width=\textwidth]{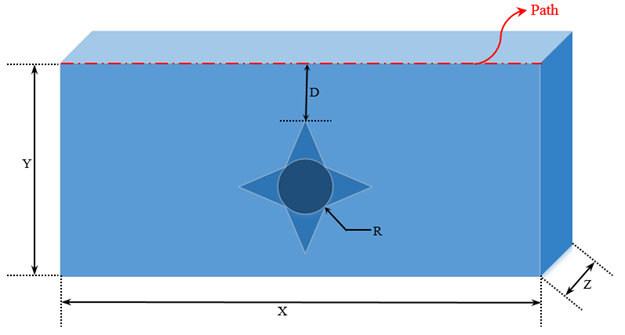}
        \caption{}
        \label{fig:starmodelfirst}
    \end{subfigure}
    \begin{subfigure}{0.5\textwidth}
        \includegraphics[width=\textwidth]{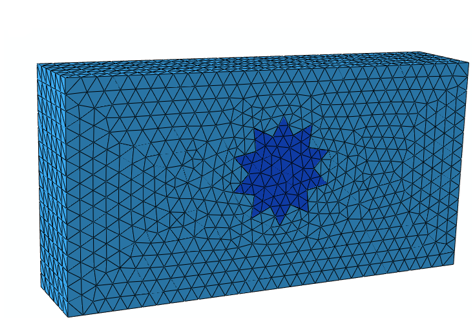}
        \caption{}
        \label{fig:starmodelsecond}
    \end{subfigure}
	\caption{Mid cross section of the brain tissue model (rectangular cuboid) in the ABAQUS environment including (a) a 4 wing star polygonal based prismatic tumor; (b) a 10 wing star polygonal based prismatic tumor.}
	\label{fig:starmodel}
\end{figure}

Figure~\ref{fig:ct} and Figure~\ref{fig:ct2} are CT scan images of two malignant brain tumors. It can be inferred that while a benign tumor has an almost smooth round shape, malignancy can be identified by the existence of invading edges. Two major morphologies of the malignant brain tumors are identified by solid lines in Figure~\ref{fig:ct} and Figure~\ref{fig:ct2}. The first morphology resembles a polygonal shape while the second one resembles a star. In the present numerical analysis, the tumor malignancy was simulated in two different scenarios:
\begin{enumerate}
    \item Tumor was considered as an n-sided polygonal based prism. Area of the polygonal base and consequently the volume of the tumor and the distance of uppermost vertex (D) were kept unchanged for all number of sides (n) (Figure~\ref{fig:prismmodel}). 
    \item Tumor was considered as a star polygonal based prism. This star was formed by an inscribed circle with constant radius of R=10mm and different corner vertices. Tumor penetration inside the nearby tissue was increased by increasing the number of corner vertices. Each group of two intersecting edges was called a wing. The number of wings was increased by keeping the star polygonal base area constant (and the tumor volume) (Figure~\ref{fig:starmodel}).
\end{enumerate}

\begin{table}
	\caption{Dimensions of simulated tissue containing a malignant tumor, all dimensions are in millimeters.}
	\centering
	\begin{tabular}{lllll}
		\toprule
		X &Y & Z &D &R \\
		\midrule
        120 &60 &25 &12 &10 \\
		\bottomrule
	\end{tabular}
	\label{tab:dim}
\end{table}


Geometrical dimensions of the sample brain tissue containing a malignant tumor is presented in Table~\ref{tab:dim}. In order to describe mechanical behavior of the brain tissue, the elasticity parameters of the brain tissue obtained by \cite{Soza_2005}) were used(Table~\ref{tab:prop}). The Young’s modulus of tumor was considered to be 10 times of the brain tissue while the Poisson’s ratio was the same \cite{Siddiqi2011}.

\begin{table}
	\caption{Elastic properties of the brain tissue.}
	\centering
	\begin{tabular}{ll}
		\toprule
        Young’s modulus (Pa) & 	Poisson’s ratio \\
        \midrule
        9210.87&	0.458344\\
        \bottomrule
	\end{tabular}
	\label{tab:prop}
\end{table}

Study of energy transport in the biological systems involves various mechanisms including conduction, advection and metabolism~\cite{Gore_2003}. In this study these phenomena were considered in a steady-state thermal analysis; a constant thermal conductivity equal to 0.6 W/m2K was considered for the brain tissue~\cite{Elwassif_2006}, blood perfusion and metabolic activities were considered by assuming a constant heat generation equal to 100000 W/m3 for the tumor. A convective heat transfer between the top tissue surface and the surrounding environment was considered with a convective heat transfer coefficient equals 20 W/m2K. The bottom surface of the brain tissue sample was assumed to have a constant temperature of 33.1\textdegree C equal to the temperature of the blood vessel which was assumed to be in contact with the tissue. Side surfaces were insulated.\\
ABAQUS software (version 6.14) was employed to perform finite element analysis (FEA) under 3D axisymmetric conditions. A compressive strain was applied to the top surface of the tissue and the temperature variation was recorded on it. In order to measure temperature variations while the tissue was loaded mechanically, the step named “COUPLED TEMP-DISPLACEMENT” was selected. A “time increment” and a “time period” should be assigned to the step for that the default values were assumed. For both rectangular cube and prism, a 4-noded tetrahedral with thermal coupling (C3D4T) element type was used.\\
Rather than thermal boundary conditions, mechanical boundary conditions were also defined. The bottom surface of the tissue was totally fixed while the top surface was loaded by a compressive strain equals to 6\% which corresponds to a 4 mm compression. Adhesion of the malignant tumor to its surrounding tissue was satisfied by applying ‘TIE’ constraint between the brain tissue and the tumor.\\
Mesh independency was examined for the tissue sample including a decagonal based prismatic tumor. Three stages of mesh refinement were performed and temperature values were measured. For three models with 21796, 29725 and 44029 elements, the computational time was 19.27, 23.44 and 29.82 seconds, respectively. The maximum relative difference between the temperature values in the models with 21796 elements and 29725 elements was less than 1\%, so the computational grid with 21796 elements was selected (Table~\ref{tab:mesh}).

\begin{table}
	\caption{Mesh independency evaluation for the sample brain tissue including a malignant tumor.}
	\centering
	\begin{tabularx}{\textwidth}{XXXXXXXX}
		\toprule
        \multicolumn{4}{c}{decagonal based prismatic tumor} &\multicolumn{4}{c}{10 wing star polygonal based prismatic tumor}\\
        \midrule
        Number of elements&	Element type&	Maximum error relative to selected model (\%)&	Computational time (sec)&		Number of elements&	Element type&	Maximum error relative to selected model (\%)&	Computational time (sec)\\
        \midrule
        21796&	C3D4T&	0&	19.27&		22362&	C3D4T&	0&	21.49\\
        29725&	C3D4T&	0.0751&	23.44&		29869&	C3D4T&	0.0665&	23.56\\
        44029&	C3D4T&	0.0382&	29.82&		43366&	C3D4T&	0.0269&	30.81\\
		\bottomrule
	\end{tabularx}
	\label{tab:mesh}
\end{table}

Similarly, mesh independency was examined for the tissue sample including the star polygonal based prismatic tumor with 10 wings. For three models with 22362, 29869 and 43366 elements, the computational time was 21.49, 23.56 and 30.81 seconds, respectively. The maximum relative error between the temperature values in the models with 22362 elements and 29869 elements was less than 1\%, so the computational mesh with 22362 elements was selected (Table~\ref{tab:mesh}).\\
The sample model of the brain tissue was numerically analyzed while a tumor with irregular borders was included as a prism with a polygonal base or a star polygonal base. The number of sides of the polygon and wings of the star were changed to the study the effects of the irregularity increase on the thermal parameters. Table 4 lists the number of sides and wings of the tumor model. A total number of 98 polygonal based prisms and 98 star polygonal based prisms were modeled and analyzed by the ABAQUS software linked with a PYTHON script to automatically change the number of sides and wings. MATLAB software (version 8.6) was used to plot the thermal outputs and extract thermal variables that would be employed as the inputs of an artificial neural network. 
\begin{table}
	\caption{Number of sides and wings of both tumor models.}
	\centering
	\begin{tabular}{lll}
		\toprule
        Initial value&	Increment&	Final value\\
        \midrule
        3&	1&	100\\
		\bottomrule
	\end{tabular}
	\label{tab:table1}
\end{table}

\subsection{Artificial neural network}
Artificial neural networks (ANNS) try to model the processing capabilities of actual human nervous systems. They are consist of numerous simple computing components, called neurons that generates an input layer, one or more hidden layers and an output layer~\cite{Rojas_1996}. ANNs are mostly used for functions approximation which may be single variable or multivariable. In order to recognize inputs patterns, suitable weights for the connections must be derived. Therefore, the network should be trained to obtain proper desired results.\\
In this study, radial base function neural network (RBF) was used that employs the radial basis functions as the activation functions~\cite{Park_1991}. The output of this network is a linear combination of neuron parameters and radial basis function of the inputs. Properties of the employed network are listed in Table~\ref{tab:ann}.

\begin{table}
	\caption{Neural network’s properties}
	\centering
	\begin{tabularx}{\textwidth}{XXXXXXXXX}
		\toprule
        Performance function&	Network initialization function&	Network derivative function&	Number of inputs&	Number of outputs&	Transfer function&	Type of neurons in 1st layer&	Type of neurons in 2nd layer&	Number of epochs\\
        \midrule
        MSE&	initlay&	defaultderiv&	10&	1&	Tansig&	Radbas&	Purelin&	100\\
		\bottomrule
	\end{tabularx}
	\label{tab:ann}
\end{table}
Performance of this network was evaluated by mean squared error (MSE) that is the average of the squares of the errors and error is the difference between the desired value of a variable and the estimated value by the network (equation~\ref{eq:mse}). The ideal performance of a neural network occurs when values of the MSE are zero. The error is defined as equation~\ref{eq:error}:

\begin{equation}\label{eq:error}
e_i=\hat x_i-x_i                                                                                           \end{equation}

Where $\hat x_i$ is the estimated value and $x_i$ is the desired value that is obtained from the ABAQUS runs in this study. Mean of the errors for all data (98 data for each tumor model) is then calculated by equation~\ref{eq:mean}:
\begin{equation}\label{eq:mean}
\mu=\frac{1}{n}\sum_{i=1}^{n} e_i                                                                          \end{equation}                                                                                                                                                
\begin{equation}\label{eq:mse}
MSE=\frac{1}{n}\sum_{i=1}^{n} e_i^2                                                                        \end{equation}

Where $n$ is the number of the network inputs. The variance is the average of the squared differences from the mean (equation~\ref{eq:sigma2}):

\begin{equation}\label{eq:sigma2}
\sigma^2=\frac{1}{n}\sum_{i=1}^{n} (e_i-\mu)^2                                                             \end{equation}
In order to have a better criterion for evaluating the network performance, it’s better to use the root mean square error (RMSE) which has the same unit as the variable being estimated (equation~\ref{eq:rmse}). The square root of the variance is known as the standard deviation (equation~\ref{eq:sigma}).
\begin{equation}\label{eq:rmse}
RMSE=\sqrt{\frac{1}{n}\sum_{i=1}^{n} e_i^2}
\end{equation}

\begin{equation}\label{eq:sigma}
\sigma=\sqrt{\frac{1}{n}\sum_{i=1}^{n} (e_i-\mu)^2}
\end{equation}

\section{Results and Discussion}
\begin{figure}
	\centering
    \begin{subfigure}{0.23\textwidth}
        \includegraphics[width=\textwidth]{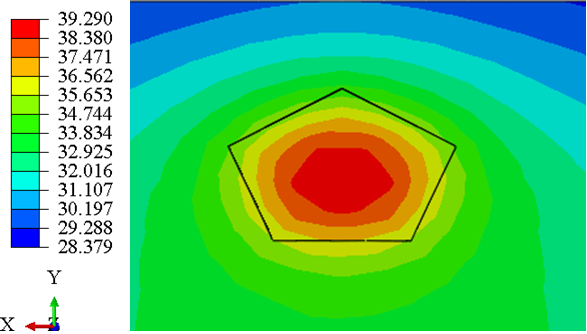}
        \caption{}
        \label{fig:contour5poly}
    \end{subfigure}
    \begin{subfigure}{0.23\textwidth}
        \includegraphics[width=\textwidth]{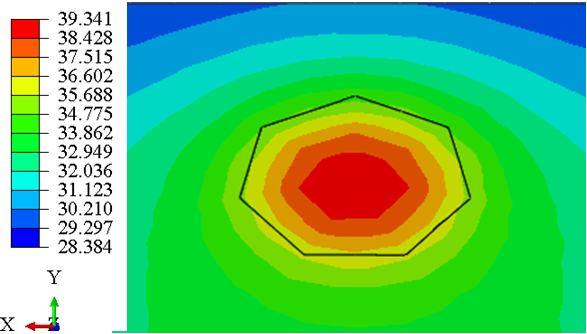}
        \caption{}
        \label{fig:contour7poly}
    \end{subfigure}
    \begin{subfigure}{0.23\textwidth}
        \includegraphics[width=\textwidth]{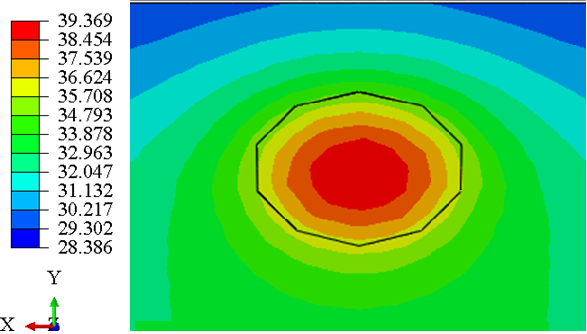}
        \caption{}
        \label{fig:contour10poly}
    \end{subfigure}
    \begin{subfigure}{0.23\textwidth}
        \includegraphics[width=\textwidth]{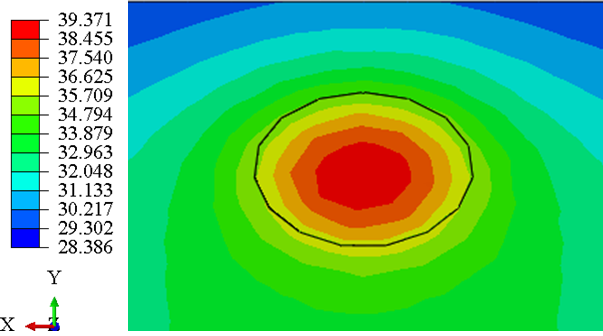}
        \caption{}
        \label{fig:contour30poly}
    \end{subfigure}
    \begin{subfigure}{0.23\textwidth}
        \includegraphics[width=\textwidth]{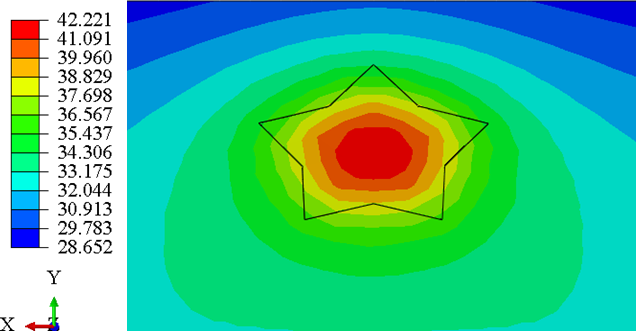}
        \caption{}
        \label{fig:contour5star}
    \end{subfigure}
    \begin{subfigure}{0.23\textwidth}
        \includegraphics[width=\textwidth]{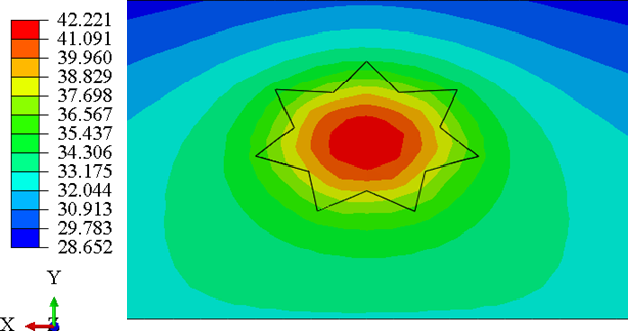}
        \caption{}
        \label{fig:contour7star}
    \end{subfigure}
    \begin{subfigure}{0.23\textwidth}
        \includegraphics[width=\textwidth]{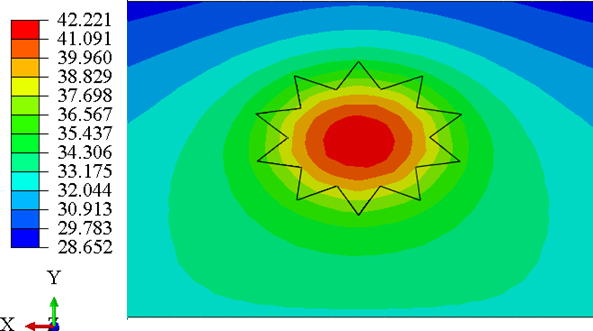}
        \caption{}
        \label{fig:contour10star}
    \end{subfigure}
    \begin{subfigure}{0.23\textwidth}
        \includegraphics[width=\textwidth]{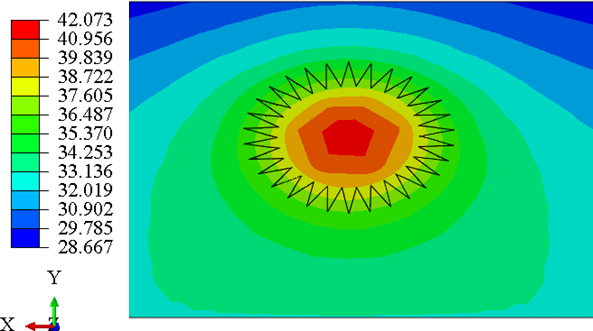}
        \caption{}
        \label{fig:contour30star}
    \end{subfigure}
	\caption{Temperature contours in the mid cross section of the brain tissue model including: a pentagonal, heptagonal, decagonal and 15 sided polygonal based prismatic tumor (top from left to right) and a 5 wing, 7 wing, 10 wing, and 30 wing star polygonal based prismatic tumor (bottom from left to right).
}
	\label{fig:contour}
\end{figure}

Figure~\ref{fig:contour} is a display of the temperature contours in the mid cross section of the model obtained from the ABAQUS runs. Tumor existence in the tissue increases the temperature in the vicinity of the tumor. At further distance from the tumor margin, the temperature reduces and the rate of reduction is not proportional to the distance from the tumor center. Temperature gradients are considerable in the tumor vicinity. It can be also inferred from Figure~\ref{fig:magplot} that the area of the tumor-affected region depends on the tumor shape to a great extent. While the tumor effect spreads smoothly over the whole tissue area for a polygonal based prismatic tumor, it has a localized distribution for a star polygon. Consequently, temperature distribution pattern can be correlated with the shape of the malignant tumor. 

Temperature variation was studied on the paths defined in Figure~\ref{fig:prismmodelfirst} and Figure 4a. and by varying the number of sides and wings of the tumor (Figure~\ref{fig:starmodelfirst}. The diagrams show that the location of the maximum temperature on the tissue surface corresponds to the location of the tumor center inside the tissue where the tumor distance is the minimum relative to the tissue surface. Moreover, while the tumor volume was kept unchanged, increasing the number of sides and wings results in the elevation of the maximum surface temperature. These achievements offer two opportunities:

\begin{enumerate}
    \item The temperature variation is indicative of the tumor existence. Therefore, temperature map can be used for the tumor detection task.
    \item For a specific malignant tumor, the temperature map can be recorded in the successive examinations. Comparison between the maps can be indicative of the malignancy progression in a period of time.
\end{enumerate}

\begin{figure}
	\centering
        \begin{subfigure}{0.8\textwidth}
        \includegraphics[width=\textwidth]{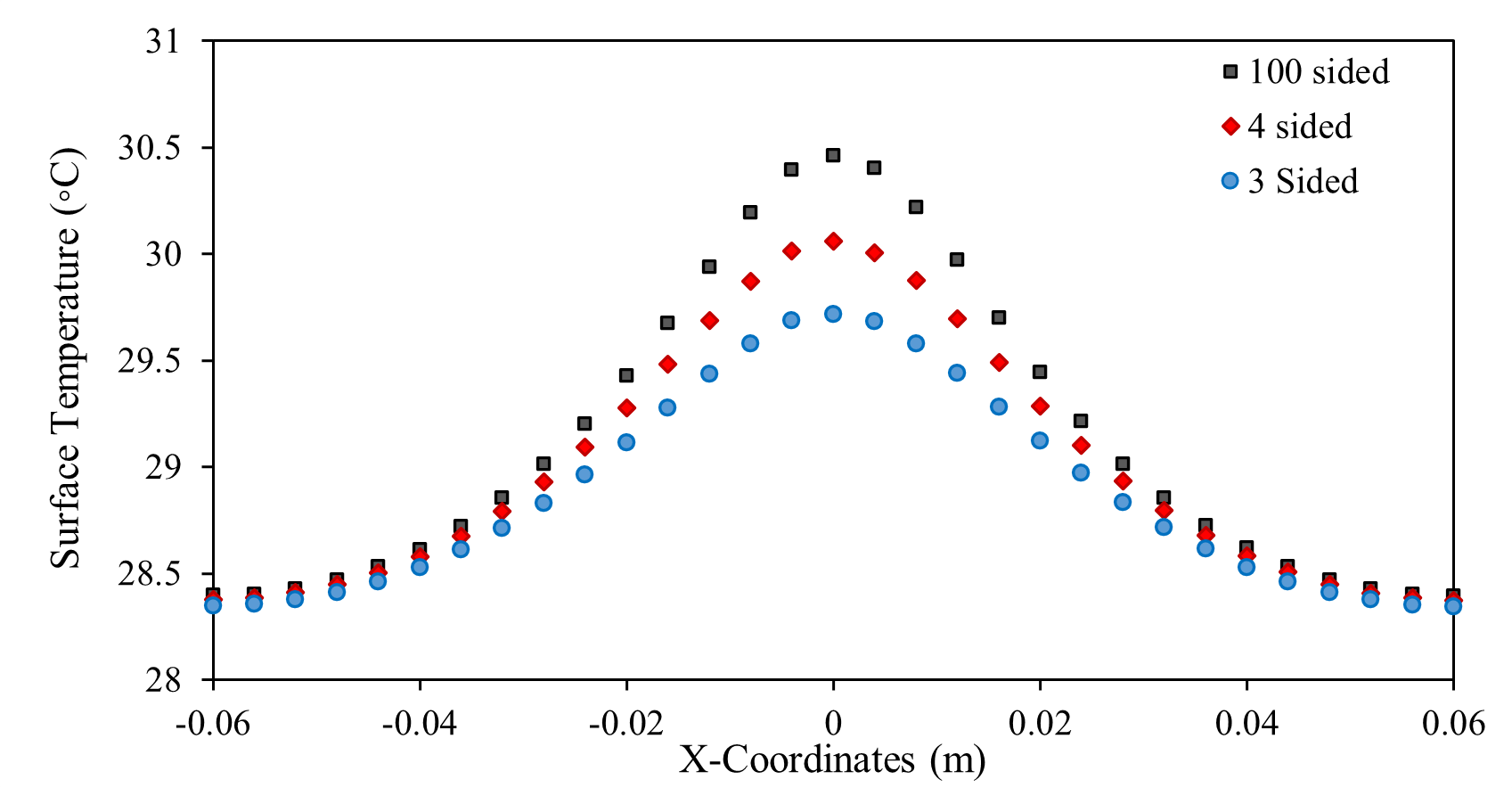}
        \caption{}
        \label{fig:tempplotfirst}
    \end{subfigure}
    \begin{subfigure}{0.8\textwidth}
        \includegraphics[width=\textwidth]{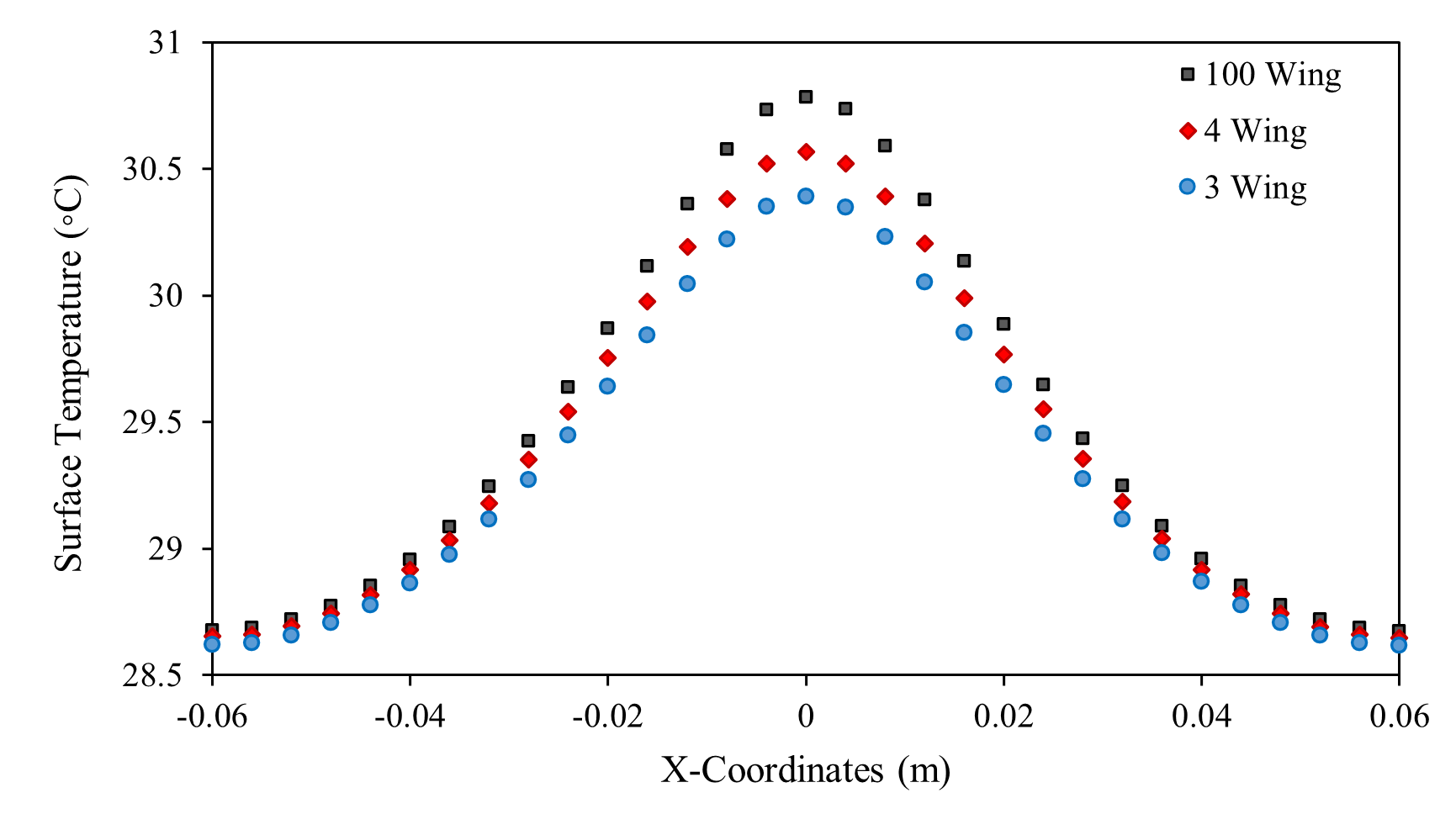}
        \caption{}
        \label{fig:tempplotsecond}
    \end{subfigure}
	\caption{Temperature distribution on a  path which passes from the center of the top tissue surface for brain tissue sample including a) polygonal based prismatic tumors; b) star polygonal based prismatic tumors.}
	\label{fig:tempplot}
\end{figure}

In Figure~\ref{fig:magplot} variation of the maximum temperature on the tissue surface is investigated by increasing the number of sides and wings. For the polygonal based tumor model, increase of the number of sides from 3 to 100 results in the elevation of the maximum surface temperature from 29.7\textdegree C to 30.5\textdegree C. For the star polygonal based tumor, by increasing the number of the wings from 3 to 100, maximum surface temperature increases from 30.3\textdegree C to 30.8\textdegree C. However, maximum temperature variations are less than 0.02\textdegree C and 0.01\textdegree C by increasing the number of sides and wings more than 20, respectively. Therefore, sensitivity of the surface temperature to the variation of the sides and wings of the tumor reduces when the sharpness of the tumor morphology increases. 

 \begin{figure}
	\centering
    \includegraphics[width=0.8\textwidth]{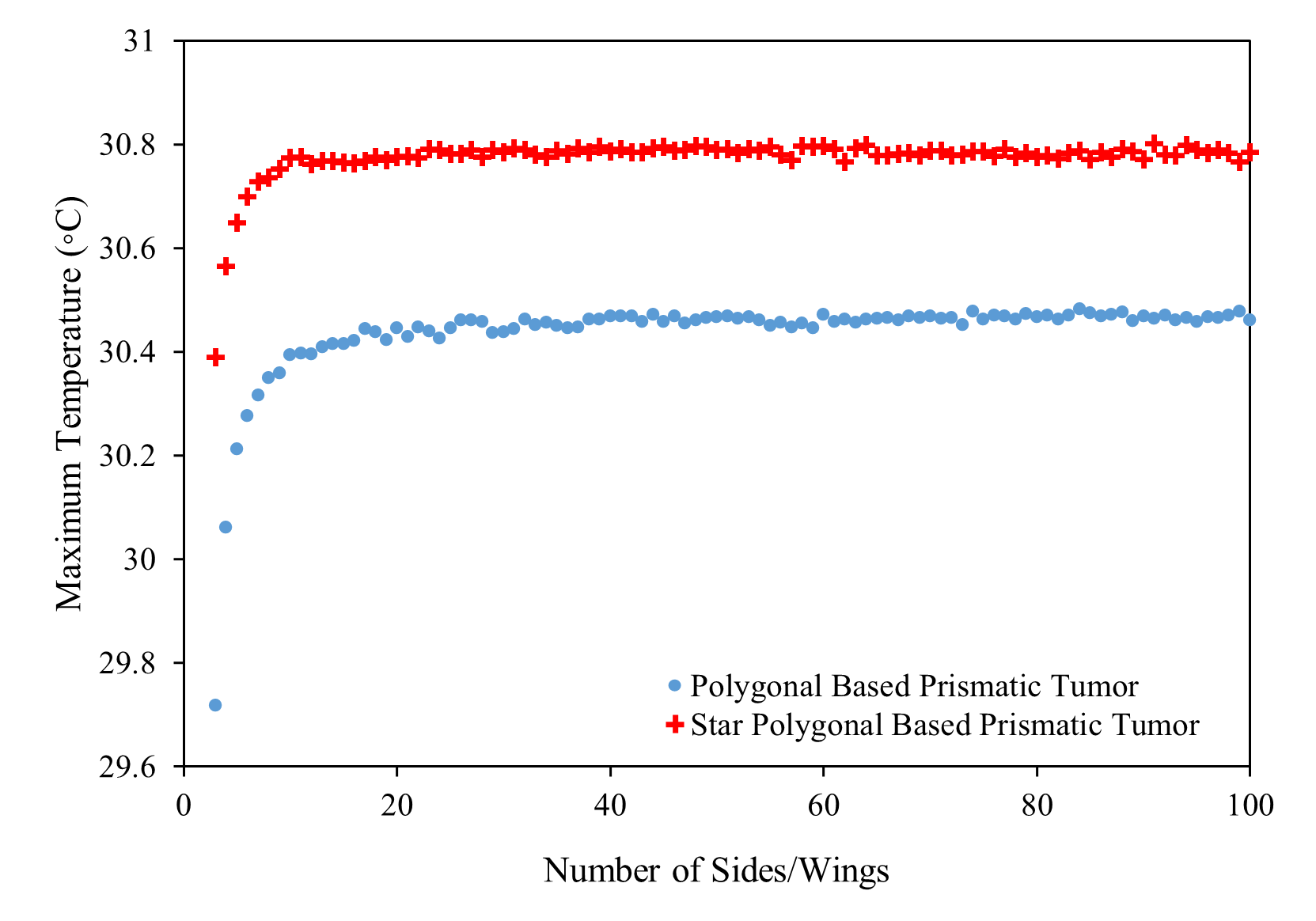}
	\caption{Correlation between the maximum surface temperature and number of sides of the polygonal based prismatic tumor and number of wings of the star polygonal based prismatic tumor.}
	\label{fig:magplot}
\end{figure}

In order to find a quantitative criterion for the tumor detection and the malignancy progression, the temperature curve on the tissue surface was interpolated by a 4th-order Fourier series (equation~\ref{eq:fourier}). Fitting error was less than 1\% for all tumor models. By fitting this curve to each set of data achieved from the ABAQUS runs, 10 coefficients were extracted for each tumor which are $a_i^\prime$ s,$b_i^\prime$ s and w in equation~\ref{eq:fourier}.  

\begin{equation}\label{eq:fourier}
T(x) = a_0 +\sum_{i=1}^4 (a_i cos wx+b_i sin wx)
\end{equation}

These coefficients were used as the inputs of a radial basis function (RBF) artificial neural network (ANN). This network (RBFNN) provided the link between the inputs that were the surface temperature interpolating function’s coefficients and outputs that were the corresponding number of sides or wings of the malignant tumor.

\subsection{Polygonal based prismatic tumor}
The RBFNN was trained to estimate number of sides by using 98 samples of the brain tissue including a polygonal based prismatic tumor with different number of sides of the polygonal base. 68 samples from the whole datasets were used randomly for training the network and 30 remaining samples were used for the testing procedure.\\

For a better comprehension of the extent and the distribution of the coefficients of equation~\ref{eq:fourier}, the mean $\bar x$ the minimum $x_{min}$, and maximum $x_{max}$ of these coefficients for the polygonal base tumor model are listed in Table~\ref{tab:var1}. The employed neural network transfer function
 was "Tansig"that takes values only between $-1$ and $1$. Therefore, the coefficients were normalized in the range of $[-1 1]$. Normalization of the dataset would prevent misinterpretation in defining the contribution of each input in the neural network.

\begin{figure}
	\centering
    \includegraphics[width=0.8\textwidth]{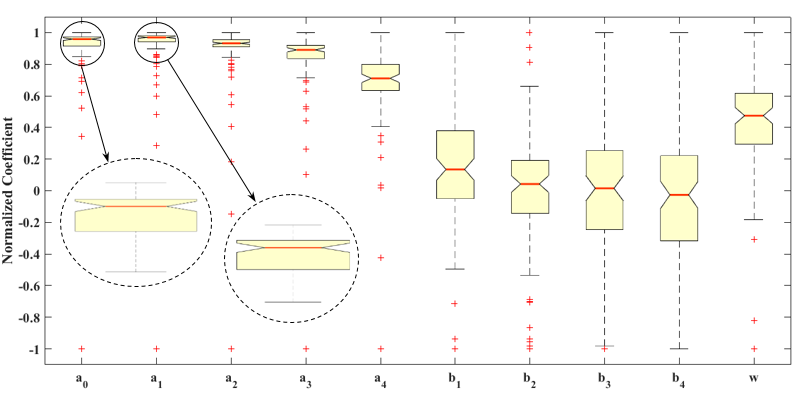}
	\caption{Box plot of the normalized coefficients for the polygon base prismatic tumor.}
	\label{fig:magplot1}
\end{figure}

\begin{table}
	\caption{Variation of the Fourier series coefficients for the polygonal based prismatic tumor.}
	\centering
	\begin{tabular}{llll}
		\toprule
        Coefficient&	$x_{min}$&	$\bar x$& $x_{max}$\\
        \midrule
        $a_0$&	28.8860&	29.1522&	29.1666\\
        $a_1$&	0.6511&	0.9437&	0.9583\\
        $a_2$&	0.1349&	0.2394&	0.2462\\
        $a_3$&	0.0326&	0.0701&	0.0737\\
        $a_4$&	0.0101&	0.0203&	0.0223\\
        $b_1$&	0.0017&	0.0042&	0.0062\\
        $b_2$&	0.000983&	0.0034&	0.0059\\
        $b_3$&	-0.000574&	0.0018&	0.0043\\
        $b_4$&	-0.000914&	0.0010&	0.0031\\
        $w$&	51.7375&	52.6726&	53.0506\\
		\bottomrule
	\end{tabular}
	\label{tab:var1}
\end{table}
The box plot of the normalized coefficients is plotted in Figure~\ref{fig:magplot1} that is a standardized way of displaying the distribution of data based on five characteristics: minimum, maximum, median, first quartile, and third quartile. In a box plot, the rectangle expands from the first quartile to the third quartile and the inner line represents the median. A segment inside the rectangle shows the median and whiskers above and below the box show the minimum and the maximum of the data. It can be inferred from Figure\ref{fig:magplot2} that $a_0$,$a_1$,$a_2$ and $a_3$ have the most compact distributions which means that these coefficients are not too sensitive to the variation of the number of sides and consequently do not have major contribution in the proposed network training. Contribution of these coefficients in the network is providing a general similarity between the patterns of temperature variations. On the contrary, $b_1$,$b_3$ and $b_4$ have the widest distributions. Moreover, based on the position of the median, a coefficient may have normal or abnormal distribution. According to the box plot, $a_2$,$a_4$, $b_3$, and $b_4$ have the most normal distributions that mean that by varying the number of sides, distribution of these coefficients is not concentrated below or above the medians. Normal distribution facilitates the network training and reduces the normal deviations. Distribution of $a_0$, $a_1$ and $a_3$ are far from being normal.

 \begin{figure}
	\centering
    \includegraphics[width=0.9\textwidth]{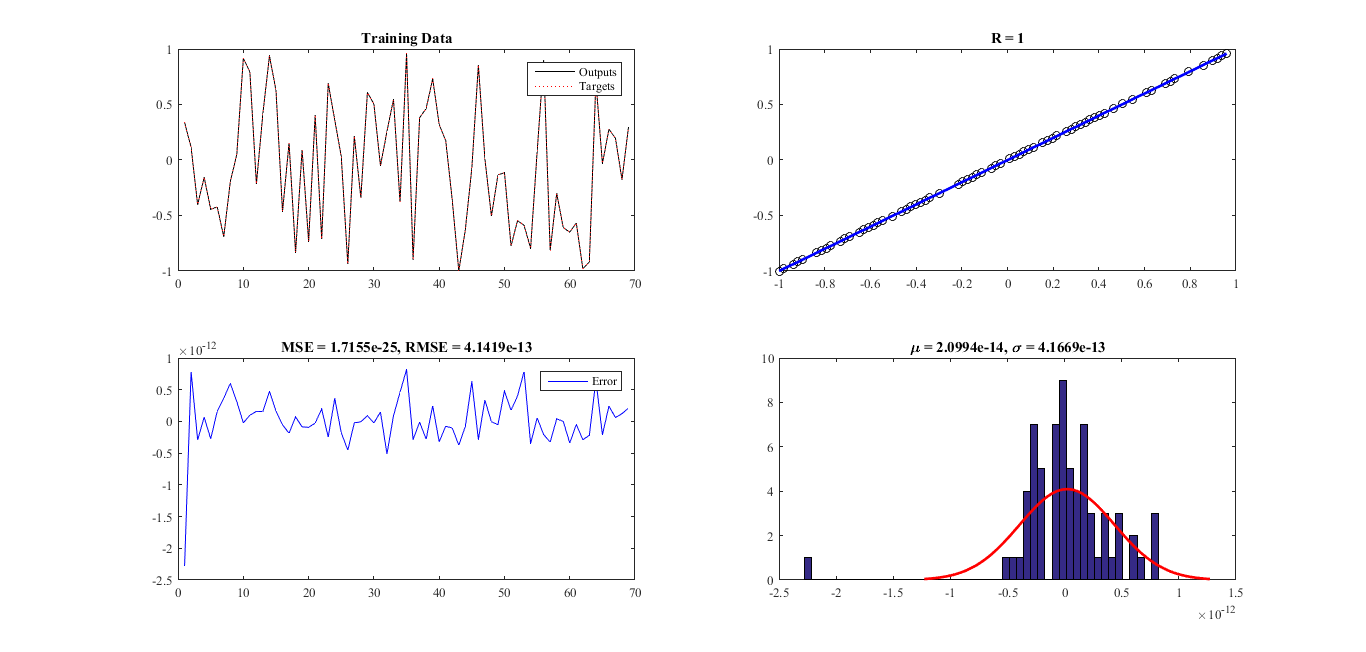}
	\caption{Performance plots of data training by the proposed RBFNN for the polygonal based prismatic tumor.}
	\label{fig:train}
\end{figure}

Performance of the proposed RBFNN in data training was evaluated by four different plots in Figure\ref{fig:train}. The top left figure is a plot of the network output that has overlap with the corresponding real values. The top right figure is a plot the linear regression of the real or desired values versus the network outputs. Slope of the fitted line by the RBFNN was close to 1 that means that the estimated values are very close to the real values. RMSE was equal to $4.142×10^{-13}$. Table~\ref{tab:train1} provides the value of the evaluating parameters for both training and testing datasets.
\begin{table}
	\caption{Evaluating parameters of the proposed RBFNN for the polygonal based prismatic tumor.}
	\centering
	\begin{tabular}{llllll}
		\toprule
        \multicolumn{3}{c}{Training dataset}& \multicolumn{3}{c}{Testing dataset}\\
        \midrule
        RMSE&	$\mu$&	$\sigma$&		RMSE&	$\mu$&	$\sigma$\\
        \midrule
        $4.142×10^{-13}$&	$2.099×10^{-14}$&	$4.167×10^{-13}$&	$3.310×10^{-12}$&	$-1.937×10^{-12}$&$2.727×10^{-12}$\\
		\bottomrule
	\end{tabular}
	\label{tab:train1}
\end{table}

\subsection{Star polygonal based prismatic tumor}
For the second tumor model, the RBFNN was trained to estimate the number of tumor wings by using 98 samples of the tissue including star polygonal based prismatic tumors. Number of the training and testing datasets were 68 and 30, respectively. Mean $\bar x$ , minimum $x_{mean}$ , and maximum $x_{max}$ of the extracted coefficients are listed in Table 8.

\begin{table}
	\caption{Variation of the Fourier series coefficients for the star polygonal based prismatic tumor.}
	\centering
	\begin{tabular}{llll}
		\toprule
        Coefficient&	$x_{min}$&	$\bar x$& $x_{max}$\\
        \midrule
        $a_1$&	0.8480&	1.0025&	1.0119\\
        $a_2$&	0.1652&	0.2044&	0.2118\\
        $a_3$&	0.0370&	0.0441&	0.0470\\
        $a_4$&	0.0087&	0.0124&	0.0155\\
        $b_1$&	0.0015&	0.0039&	0.0060\\
        $b_2$&	0.00019&	0.0030&	0.0055\\
        $b_3$&	-0.0015&	0.0012&	0.0043\\
        $b_4$&	-0.0019&	0.0007&	0.0044\\
        $w$&	51.2349&	51.6295&	52.1607\\
		\bottomrule
	\end{tabular}
	\label{tab:var2}
\end{table}

The box plot of the normalized coefficients is plotted in Figure 10. Similar to the polygonal based prismatic tumor, the box plot shows that $a_0$,$a_1$,and $a_2$ have the most compact distributions and $a_4$, $b_1$ and $b_2$ have the most normal distributions.
 \begin{figure}
	\centering
    \includegraphics[width=0.8\textwidth]{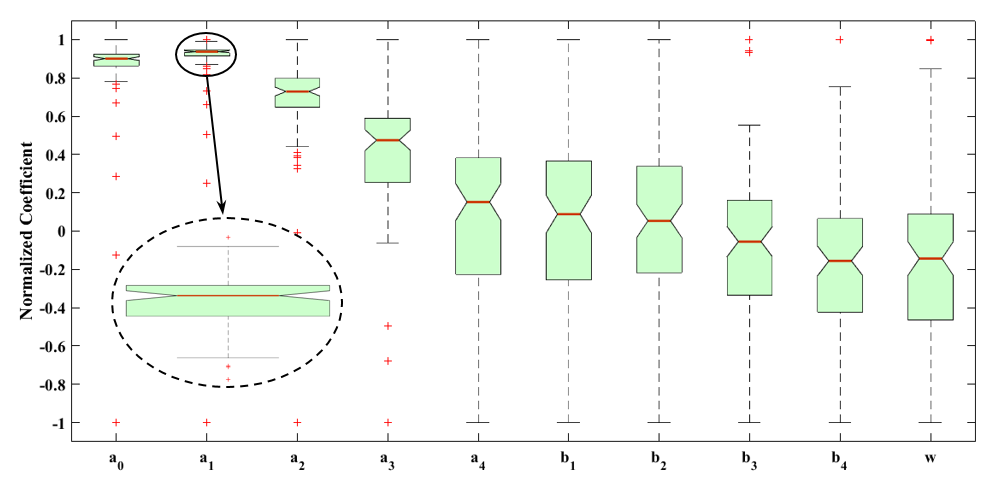}
	\caption{Box plot of the normalized coefficients for the star-shaped base prismatic tumor.}
	\label{fig:magplot2}
\end{figure}

Performance of the RBFNN was evaluated by the prescribed parameters for both training and testing datasets (Table\ref{tab:train2}). For both malignant tumor models, R-value was close to 1 and insignificant values of RMSE are indicative of perfect fit of the designed network to the numerical data. RMSE, errors mean ($\mu$) and standard deviation ($\sigma$) are almost zero.

\begin{table}
	\caption{Evaluating parameters of the proposed RBFNN for the star polygonal based prismatic tumor.}
	\centering
	\begin{tabular}{llllll}
		\toprule
        \multicolumn{3}{c}{Training dataset}& \multicolumn{3}{c}{Testing dataset}\\
        \midrule
        RMSE&	$\mu$&	$\sigma$&		RMSE&	$\mu$&	$\sigma$\\
        \midrule
        $7.028×10^{-13}$&	$-2.099×10^{-13}$&	$6.742×10^{-13}$&		$2.846×10^{-12}$&	$1.987×10^{-12}$&$2.072×10^{-12}$\\
		\bottomrule
	\end{tabular}
	\label{tab:train2}
\end{table}

\section{Conclusion}
In the present study, a morphologically malignant tumor was simulated in a cuboid sample of the brain tissue and thermal effect of the tumor on the surrounding tissue was investigated. Considering CT-scan images, morphology of the malignant tumor was defined by two major scenarios; polygonal based prismatic tumor and star polygonal based prismatic tumor. Main characteristic of both morphologies is having corner vertices and multiple edges. The tumor was considered as a biological heat source in the tissue and the temperature map on the tissue surface was obtained. An interpolating function was fitted to the temperature map and ten distinct variables were extracted. The tumor growth and malignancy progression was linked to the increase of the corner vertices of the tumor. Subsequently, 98 polygonal based prismatic tumors with different number of sides and 98 star polygonal based prismatic tumors with different number of wings were modeled and thermally analyzed. The aforementioned variables were extracted for all tumor models. Numerical results showed that the temperature of the normal tissue is affected by the tumor existence and the pattern of temperature variation has agreement with the tumor morphology. Moreover, the extracted thermal variables for all tumor models were used as the inputs of a radial base function neural network (RBFNN) and the number of sides and wings were estimated. The RBFNN analysis offered that the proposed method has the potential to be employed as a quantitative tool for measuring the malignancy progression over time.

\bibliographystyle{unsrtnat}
\bibliography{references}  %%% Uncomment this line and comment out the ``thebibliography'' section below to use the external .bib file (using bibtex) .





% \end{thebibliography}

\end{document}